\documentclass[10pt,twocolumn,letterpaper]{article}

\usepackage{iccv}
\usepackage{times}
\usepackage{epsfig}
\usepackage{graphicx}
\usepackage{amsmath}
\usepackage{amssymb}

\usepackage[pagebackref=true,breaklinks=true,letterpaper=true,colorlinks,bookmarks=false]{hyperref}

\iccvfinalcopy 

\def\iccvPaperID{1} 

\newcommand\blfootnote[1]{%
  \begingroup
  \renewcommand\thefootnote{}\footnote{#1}%
  \addtocounter{footnote}{-1}%
  \endgroup
}

\ificcvfinal\fi
\begin{document}

\title{ReActNet: Temporal Localization of Repetitive Activities\\ in Real-World Videos} 

\author{Giorgos Karvounas$^{1,2}$, Iason Oikonomidis$^{1}$, Antonis Argyros$^{1,2}$\\
${^1}${Institute of Computer Science, FORTH, Greece}\\
${^2}${Computer Science Department, University Of Crete, Greece}\\
{\tt\small \{gkarv, oikonom, argyros\}@ics.forth.gr}
}

\maketitle

\begin{abstract}
We address the problem of temporal localization of repetitive activities in a video, i.e., the problem of identifying all segments of a video that contain some sort of repetitive or periodic motion. To do so, the proposed method represents a video by the matrix of pairwise frame distances. These distances are computed on frame representations obtained with a convolutional neural network. On top of this representation, we design, implement and evaluate ReActNet, a lightweight convolutional neural network that classifies a given frame as belonging (or not) to a repetitive video segment. An important property of the employed representation is that it can handle repetitive segments of arbitrary number and duration. Furthermore, the proposed training process requires a relatively small number of annotated videos.  
Our method raises several of the limiting assumptions of existing approaches regarding the contents of the video and the types of the observed repetitive activities. Experimental results on recent, publicly available datasets validate our design choices, verify the generalization potential of ReActNet and demonstrate its superior performance in comparison to the current state of the art. 

\end{abstract}

\section{Introduction}

Repetitive\blfootnote{Accepted for presentation as a regular paper in the Intelligent Short Video workshop, organized in conjunction with ICCV'19} patterns are ubiquitous in both natural and man-made environments. 
Common human motions and activities such as walking, running, hand waving, breathing, etc, form temporally repetitive patterns.
Detecting such patterns is both effortless and useful for humans~\cite{Johansson1973} 
who use visual, aural and tactile signals as the primary sources of sensory information to solve this task. 

\begin{figure}[t]
\centering
\includegraphics[width=\columnwidth]{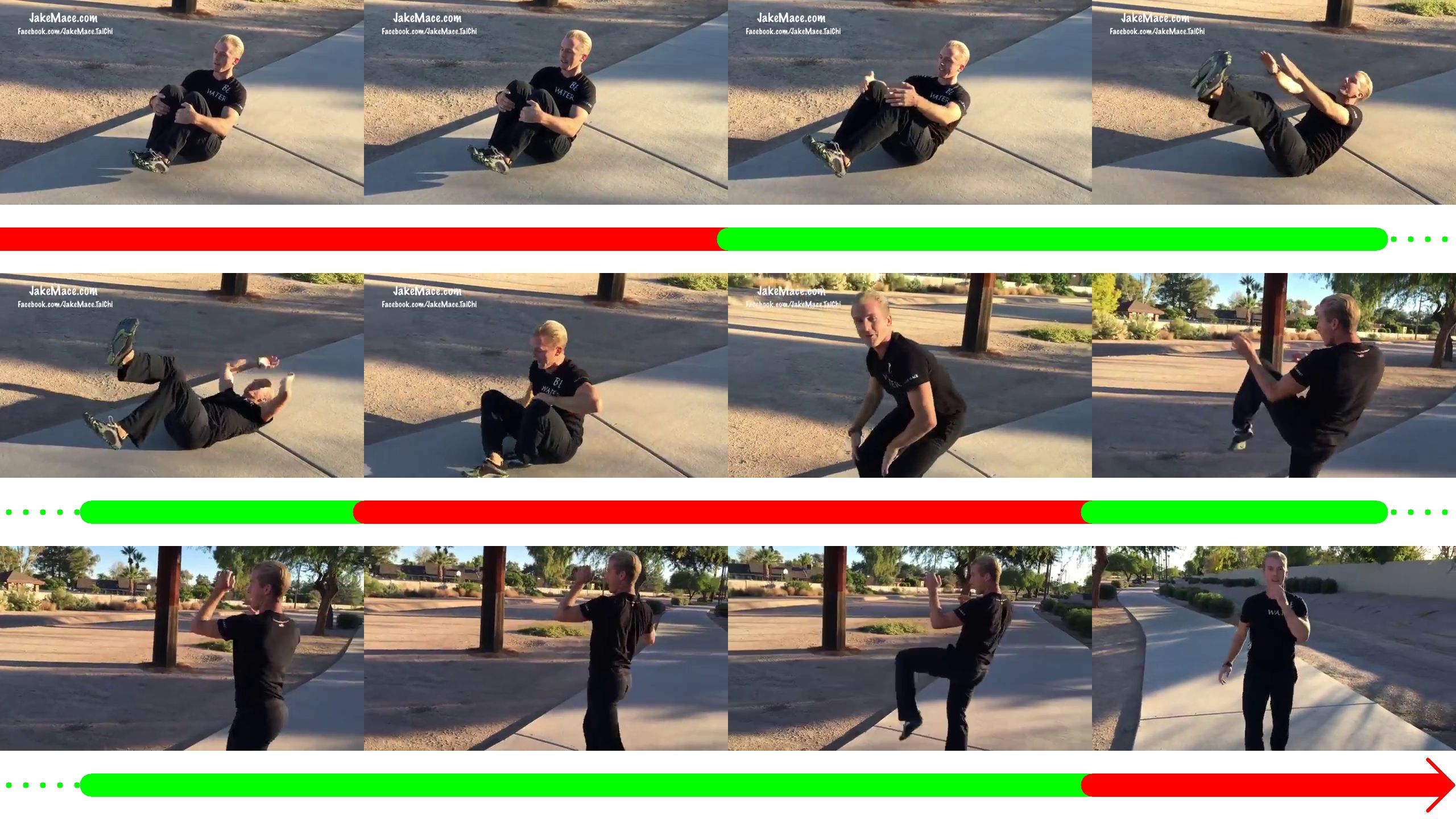}
\caption{We address the problem of classifying parts of an input video as belonging to repetitive activities or not. Still frames of an example video are shown, with the time running from left to right and from top to bottom. In the video, a man starts by talking to the camera (non-repetitive activity), followed by doing crunches (repetitive activity). Then he stands up (non-repetitive), performs some kicks (repetitive) and finally again talks to the camera (non-repetitive). The output of the proposed method is color-coded as red for non-repetitive parts, and green for repetitive ones.
\label{fig:concept}}
\end{figure}

This work deals with the problem of {\em temporal localization of repetitive activities}, that is, the identification of all repetitive segments in a video. More specifically, given an input video, our goal is to identify all the frames of the video where a repetitive, periodic motion is observed. Consider, as an example, the scenario shown in Figure~\ref{fig:concept}.
In this example, a man starts by talking to the camera (non-repetitive activity), followed by doing crunches (repetitive activity), then stands up (non-repetitive), performs some kicks (repetitive), and finally again talks to the camera. The desired output is color-coded as red for non-repetitive video segments, and green for repetitive ones. Given this identification of repetitive segments, a potential next goal is {\em repetition counting}, also called {\em periodicity characterization} in the relevant literature: the estimation of the length of the period of the repetitive motion in each segment. Repetition counting is out of the scope of this work. However, a solution to the problem of localizing repetitive segments simplifies the problem of repetition counting.

Repetitive activity detection and repetition counting using visual input form a category of problems that has been widely studied~\cite{Liu1998,Vlachos2005}, and is still of significant interest to the community of computer vision \cite{Panagiotakis,Runia2018a}. The relevant methods need to address a number of challenges. To begin with, the amount of data contained in a regular video stream is big, even for the standards of modern hardware. Further complicating factors include camera motion, sensor noise and the non-stationarity of the observed motion. 
%
Real-world signals deviate from being stationary and truly periodic. 
Repetitions of an activity may differ in their duration (thus, the period of the activity is not constant) and/or in their actual execution.
Therefore, successful repetition detection methods that deal with these types of non-stationarity have to address significant theoretical challenges. 
%
From a practical point of view, many real-world applications such as human action and motion detection, analysis and classification~\cite{Lu2004,Goldenberg2005,Albu2008,Ran2007}, 3D reconstruction~\cite{Belongie2006}, camera calibration~\cite{Huang}, medical assessment \cite{ID1} and rehabilitation, industrial inspection with minimal prior knowledge~\cite{Karvounas}, etc, can benefit from the development of robust solutions to periodicity-related problems.



\section{Related Work}
The detection and the analysis of repetitive patterns have received the attention of researchers in various computer science fields~\cite{Agrawal1995,polana1997detection,Elfeky2005,Pogalin2008} for, e.g., the detection of repeated actions within a video~\cite{polana1997detection,Pogalin2008}, of repeated patterns in a database~\cite{Agrawal1995,Elfeky2005,Exarchos2008}, etc.

\vspace*{0.2cm}
{\noindent \bf Repetition detection and characterization:}
The problems of repetition detection and repetition counting can be defined in the temporal and spatial domains~\cite{Cutler2000} but also in several others that involve either 1D signals~\cite{Lu2004,Elfeky2005,Burghouts2006} or multidimensional signal representations.
Naturally, Fourier transform and, more generally, spectral analysis tools have been employed by many methods~\cite{polana1997detection,Cutler2000,ID1}. This  yields satisfactory results in clean, 1D signals, but fails in the presence of large amounts of noise or other sources of signal corruption that invalidate the assumption of signal stationarity. Special care must then be taken to preprocess the input appropriately so that the resulting signal(s) meet the requirements imposed by the selected spectrum analysis tool.

The work by Lu et al.~\cite{Lu2004} tackles the problem of segmenting and classifying repetitive motion events using visual input. The method decomposes the input video into motion primitives using a multidimensional signal segmentation algorithm and then uses these primitives to segment and classify repetitive motion.
Burghouts and Geusebroek~\cite{Burghouts2006} propose a method to detect and characterize periodic motion. A spatiotemporal filtering approach is adopted, yielding localization of the observed periodic events. 
The method of Tong et al.~\cite{Tong2005} operates on local motion (optical flow) information~\cite{Black1996} computed in videos. Repetition detection and characterization is performed by estimating statistics of the autocorrelation computed over this motion representation. Albu et al.~\cite{Albu2008} use silhouettes as the main visual cue. They focus on human silhouettes to obtain 1D signals representing motion trajectories of body parts. These signals are then processed to detect their repetitve parts, with a final step of the method fusing these detections into a single coherent estimation of the repetitive parts of the input. The authors show experimental results of their method on repetitive activities such as aerobic exercises and walking.

Briassouli and Ahuja~\cite{Briassouli2007} propose a method to extract multiple periodic motions in a video sequence. They propose the use of Short Term Fourier Transform (STFT) on the volume of the input video. This approach allows for the simultaneous detection and characterization of repetitive activities.
Pogalin et al.~\cite{Pogalin2008} define $10$ different classes of repetitive motions, and build a system to detect and classify them. To do so, they begin by tracking all the objects in the input stream. This is followed by probabilistic PCA of the resulting tracks, and spectral analysis for detecting and characterizing periodic motion. Azy and Ahuja~\cite{Azy2008} propose a method to detect and segment objects that move in a repetitive manner using a maximum likelihood estimation of the period to characterize the motion. Image segmentation is performed using correlation of image segments over the estimated period. Taking the inverse approach, Gaojian Li et al.~\cite{Li2012} first localize the target object in a region of interest and then characterize its periodic motion. Motions with mild non-stationary components are shown to be handled effectively.
Karvounas et al.~\cite{Karvounas} detect and characterize the periodic part of an input video by formulating the task as an optimization problem. This work assumes that only one periodic activity exists per video. In the case that this assumption does not hold, only one of the periodic activities will be detected, or the method may completely fail to detect and characterize a periodic action. Panagiotakis et al.~\cite{Panagiotakis} treats the detection of multiple periodic actions as a problem of video co-segmentation. More specifically, they capitalize on a method~\cite{panagPR} that co-segments all common actions of two videos, in an unsupervised manner: by co-segmenting actions in a single video, the proposed work manages to identify repeated motions.

\vspace*{0.2cm}
{\noindent \bf Exploiting periodicity:}
The cue of repetitive motion is very strong and can be used to detect events in a video~\cite{Lu2004,Laptev2005}. In a work demonstrating the power of the repetition cue using visual input~\cite{Sarel2005}, Sarel and Irani show that it is possible to separate two superimposed layers in a video, under the assumption that one of the layers exhibits repetitive motion dynamics. As an example, a man performing jumping jacks is filmed through a glass door that reflects moving people on the other side of the door. The proposed method separates correctly the two scenes by assuming that repetitive motion occurs in one of the two layers. Another work where repetition serves as a strong cue to detect and segment motions is presented by Laptev et al.~\cite{Laptev2005}. In that work, the authors start by observing that corresponding time instances in successive repetitions of a periodic activity serve as approximate stereo pairs. This observation is then exploited to segment and characterize repetitive activities using space-time correspondences across different repetitions of the observed motion. 
Goldenberg et al.~\cite{Goldenberg2005} propose the use of the sequence of silhouettes of a periodically moving object such as a walking dog as a basis for the representation and analysis of the motion towards action classification. The authors propose the eigen-decomposition of this set of silhouettes as a means to extract an intermediate representation. The authors proceed to show how this representation can efficiently solve the task of object and action classification. 
Xiu Li et al.~\cite{Li} exploit periodicity to develop a method for non-rigid structure from motion. The input to the method is a monocular video of a non-rigid object undergoing a possibly repetitive dynamic motion. The authors observe that many deforming shapes tend to repeat themselves in time, a property termed ``shape recurrency''. Based on this property, the authors apply standard, rigid structure-from-motion tools on this problem. 

\vspace*{0.2cm}
{\noindent \bf Repetition counting:}
Levy and Wolf~\cite{Levy2015} propose a method that tackles the problem of repetition counting. They use  a convolutional neural network on windows of the input stream to detect and characterize repetitive activities. Based on this output, they proceed to estimate the number of repetitions in the input video stream. The recent method by Runia et al.~\cite{Runia2018a} for repetition counting analyzes the different possible viewpoints of a periodic motion with respect to the dominant orientation of motion in 3D. Then, the method selects the viewpoint with the best signal, estimating the length and period of motion. This, in turn, yields an estimation of the repetition count, the end goal of this work. This method only tackles the problem of counting the repetitions of a single action, so it cannot handle two or more different periodic activities in a video.


\vspace*{0.2cm}
{\noindent \bf Our contribution:}
The overview of the relevant literature indicates that, with the exception of~\cite{Panagiotakis}, no method can tackle the problem of detecting the repetitive segments of a real world input video in its generality. Existing methods make restrictive assumptions regarding the stationarity of the observed signals. For example, they consider strictly periodic signals of almost constant period and very similar execution of actions in the different repetitions. Others, assume that a single periodic activity is executed in a video and, therefore, cannot detect several of them. 
Finally, others are tied to specific (and thus, limiting) settings as, for example, that the camera observing the scene is static. Others operate on representations of the human body, thus they may deal only with repetitive patterns of human motion. In our work, we present a deep learning method that overcomes all aforementioned limitations. 

We first employ a generic video-content representation on which we show that temporal localization of repetitive activities can be learned. We also propose ReActNet, a lightweight deep learning architecture to solve this problem.
We show that, based on the employed representation, ReActNet learns to localize repetition on the basis of a relatively small training dataset. 
Finally, we  evaluate ReActNet quantitatively on existing public datasets and in comparison with existing approaches. The experimental results show that the proposed approach outperforms the state of the art.


\begin{figure}[t]
\centering
\includegraphics[width=0.7\columnwidth]{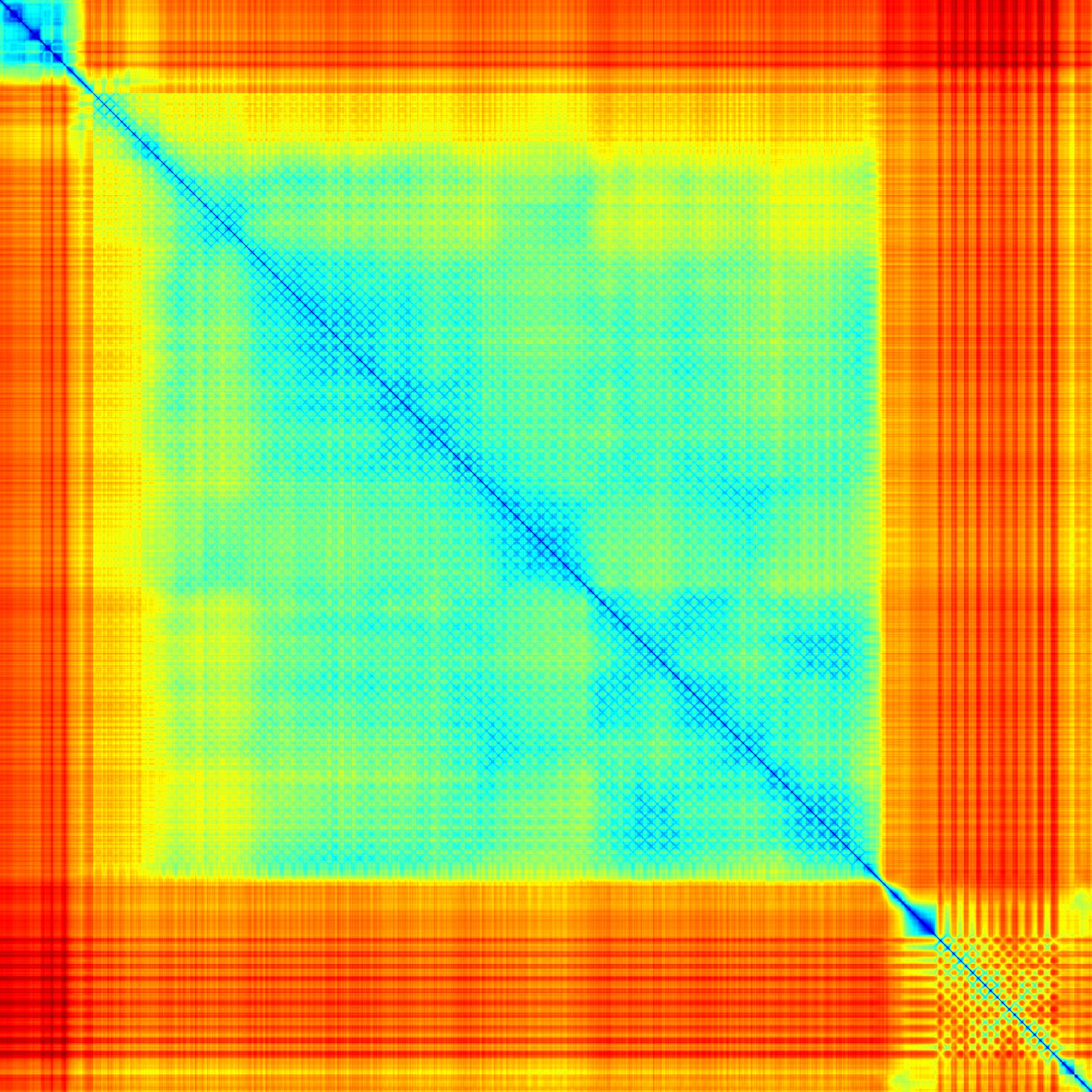}
\includegraphics[width=0.7\columnwidth,height=1cm]{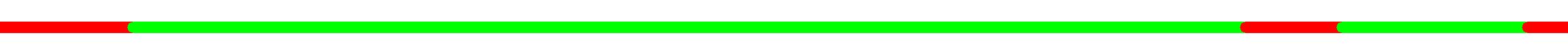} 
\caption{Top: The distance matrix $M$ of a video that contains two repetitive activities. Warm (cold) colors indicate larger (smaller) distance between frame descriptors. Bottom: 1D ground truth annotation (red: non-repetitive, green: repetitive).
\label{fig:DM}}
\end{figure}

\section{Localizing Repetitive Segments}
Given an input video, each frame is represented as a vector of deep features that encode high-level information for that frame.
Based on this frame representation, we compute a matrix $M$ of the pairwise distances of all video frames. The existence of repetitive activities in a video  gives rise to distance matrices with a particular block diagonal structure. Based on this observation, we proceed to train a neural network that classifies square blocks on the main diagonal of $M$ into two classes (repetitive / non-repetitive). Figure~\ref{fig:DM} gives an example of a distance matrix of a video showing two repetitive activities among other, non-repetitive parts.

\subsection{Data Representation}
We approach the problem of temporally localizing repetitive activities building upon the idea of a distance matrix which has been adopted also by other works in video analysis~\cite{Cutler2000, Karvounas, papoutsakis2017a, Panagiotakis}. Specifically, assuming that the input is a video of $N$ frames, we construct an $N \times N$ matrix $M$. Each entry $m_{ij}$ of $M$ quantifies the distance between frames $i$ and $j$. Assuming that some temporally repetitive pattern is present in the observed video, the matrix exhibits particular structures. For example, it may contain sub-diagonals that are parallel to the main diagonal with low distance values, capturing the periodic nature of the observations.


An intermediate representation for each frame of the input video is useful for the task of computing the entries of matrix $M$.
The representation of a frame can, in principle, be anything that captures visual content, disentangling it from signal-specific and repetition-irrelevant details. Panagiotakis et al.~\cite{Panagiotakis} opted for the use of tracklets, short sequences of optical flow trajectories. This approach can successfully separate appearance from motion, but is prone to tracking failures. In this work, we propose the use of features that are computed by a deep convolutional neural network. 
Specifically, the activation features of the VGG19 network~\cite{Krizhevsky} at the $15$th layer are computed for each frame of the input video. This particular layer has been selected as it strikes a good balance between the required level of feature abstraction and the need to keep features in spatial relation with the input. Then, for each pair $i$ and $j$ of frames of the video, the Euclidean distance of their precomputed feature vectors is computed as the value $m_{ij}$ of $M$. 
An example of such a matrix is visualized in Figure~\ref{fig:DM}.

\subsection{Ground Truth Annotation}
\label{sec:annotation}
Having computed the distance matrix $M$ based on the input video, our next task is to denote each frame of the video as part of a repetitive activity or not.
We assume that there is ground truth annotation for a number of input videos. Specifically, the annotation for a video is a list of pairs of frames of the form $(s_i, e_i)$, where $s_i$ is the frame count of the start of the $i$-th repetitive segment, and $e_i$ is the ending frame of that segment.

\begin{figure}[t]
\centering
\includegraphics[width=.48\columnwidth]{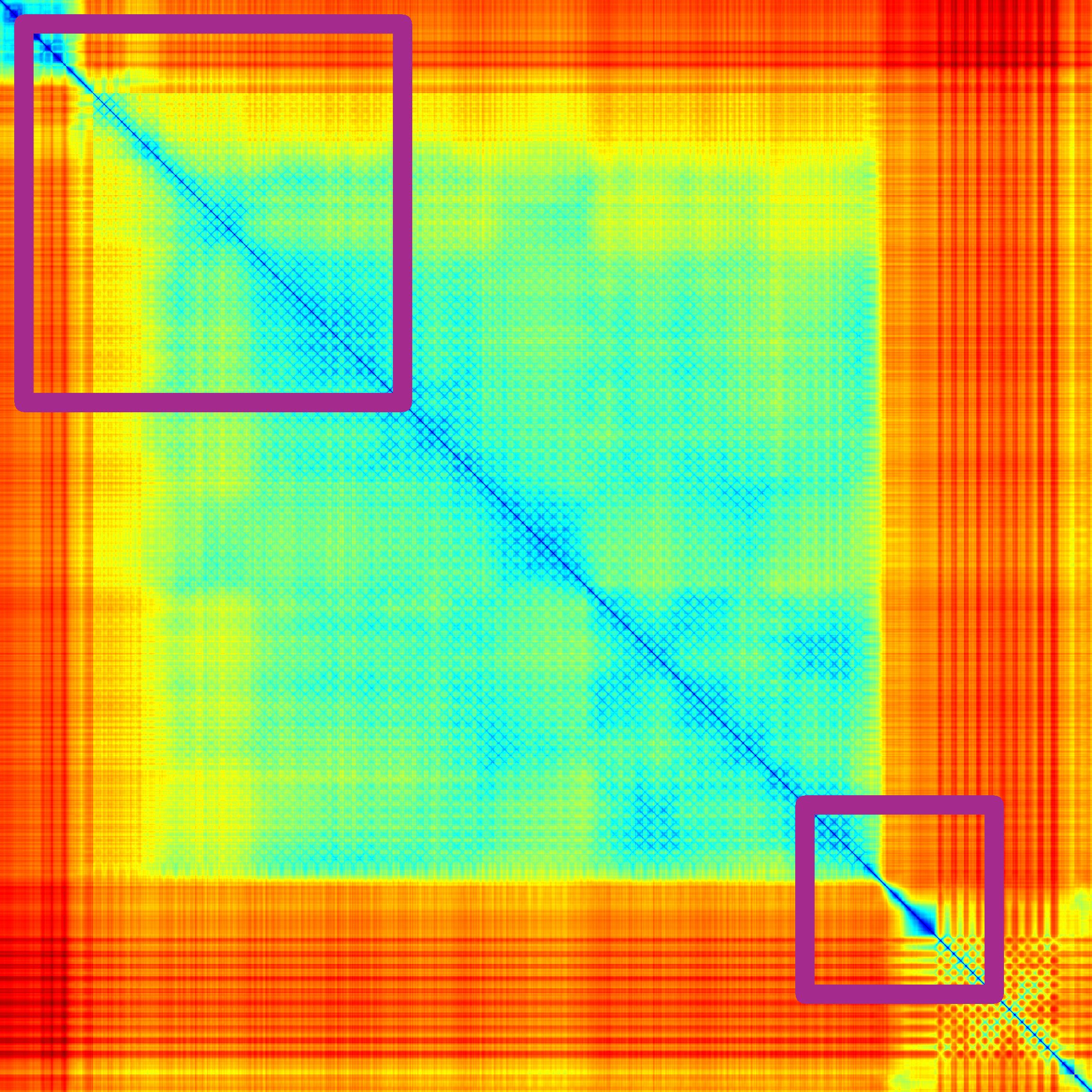}
\includegraphics[width=.48\columnwidth]{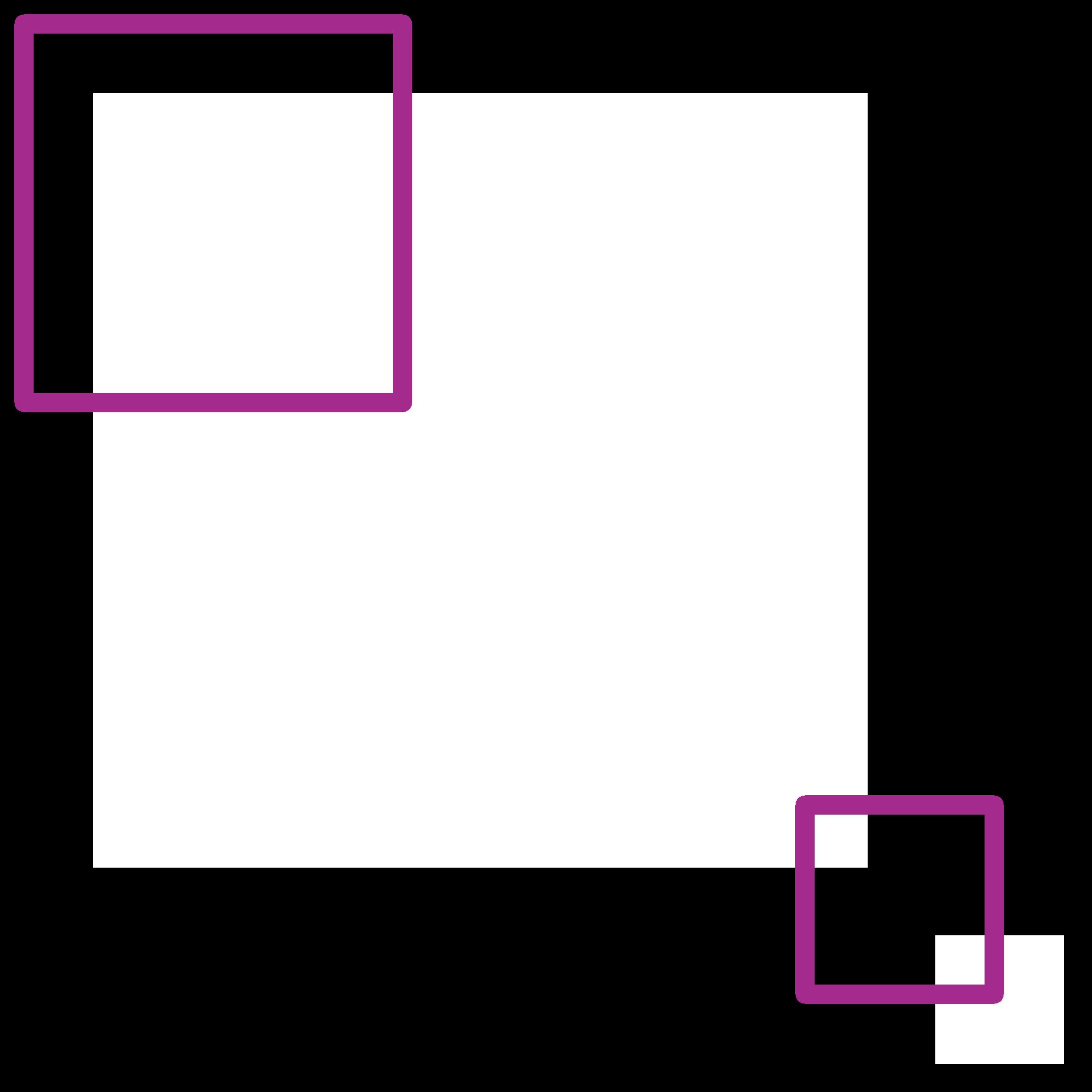}
\caption{A distance matrix $M$ (left) and the corresponding binary ground truth matrix $A$ (right). Two different sub-blocks are also shown in purple color (see text for details).
\label{fig:GT_subblocks}}
\end{figure}

For a number of reasons, it is not practical to train a neural network directly on the ground truth represented in the form of repetitive segments $(s_i, e_i)$.
Firstly, the input size of the image sequence (and of a repetitive part of it) can be arbitrarily large. This would make the resulting network prohibitively large for practical use on current hardware. A solution to the arbitrary input size is the use of recurrent neural networks. We don't adopt this design choice because of the additional complexity to fine-tune a recurrent neural network.
Additionally, this strategy has no clear way of representing multiple repetitive segments that may co-exist in a video, occurring one after the other.
To address these issues, we adopt a representation that is efficient with respect to input size, and represents the target output in a way that is compatible with a fully convolutional architecture.

\subsection{Learnable Repetition Representation}
To tackle the issues above, we resort to a binary, square matrix $A$ with the same dimensions as $M$. We adopt the convention that a value $a_{ij}$ of $A$ is equal to $1$ if both frames $i$ and $j$  belong to the same repetitive segment. Otherwise, $a_{ij} = 0$. 
Figure~\ref{fig:GT_subblocks} shows a distance matrix $M$ (left) and the corresponding binary matrix $A$ (right).
Given the annotation information as described is Section~\ref{sec:annotation}, it is straightforward to compute the respective matrix $A$.

Representing the input video content using the matrix $M$, and the ground truth annotation as the matrix $A$, we propose to train a network on fixed-size sub-blocks of these matrices.
Since the information regarding repetitions is encoded around the main diagonal of the matrix $M$, any square block of the matrix $M$ that is centered on the main diagonal of $M$ can be used as a training sample. The corresponding block of the matrix $A$ can then serve as the ground truth annotation for this training sample.

Given this information, we proceed to train a deep neural network on such training samples.
Overall, the training samples are corresponding pairs of sub-blocks of the matrix $M$ and $A$.
It is fine to select sub-blocks of $M$ that have large overlap. 
This process is repeated for all matrices $M_i$ coming from input videos $V_i$ in the training set, resulting in thousands of training sub-blocks given a few input videos.

Figure~\ref{fig:GT_subblocks} shows two example training sub-blocks (purple squares) superimposed on a distance matrix $M$ (left) and on the respective ground truth annotation matrix $A$ (right).

\begin{figure*}[t]
\centering
\includegraphics[width=0.6\textwidth]{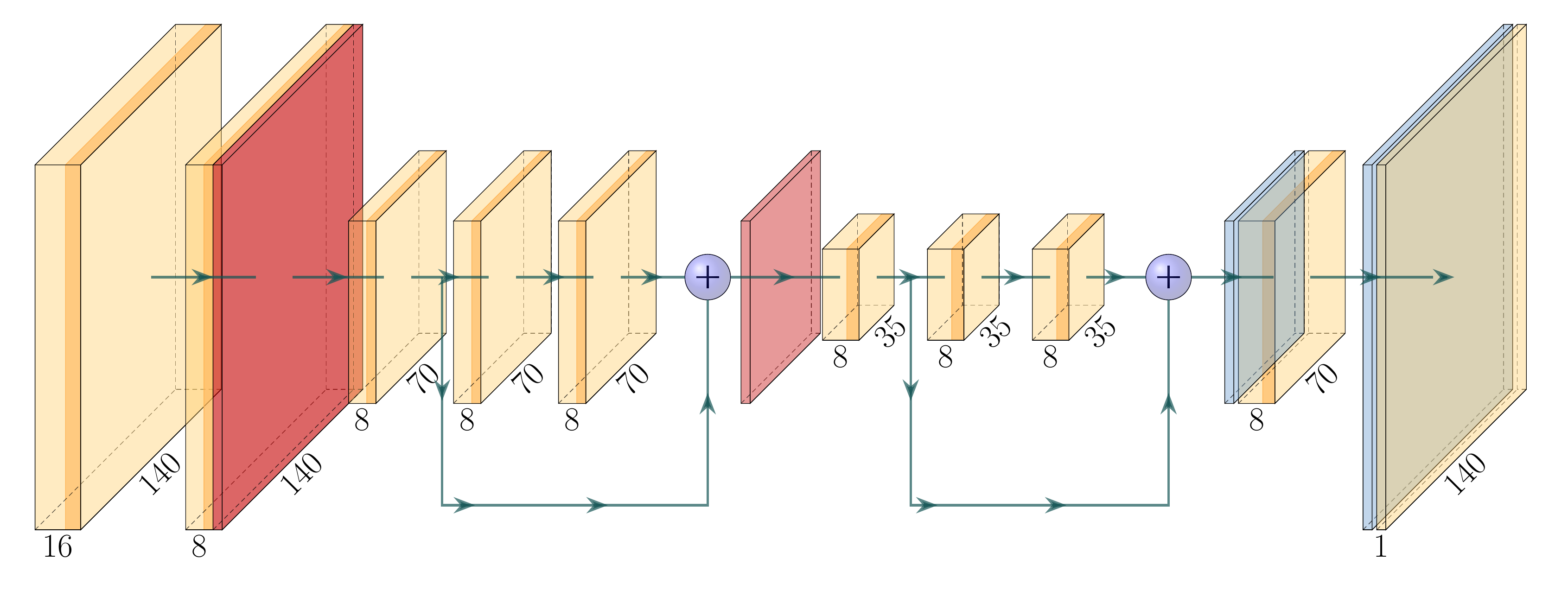} 
\caption{The building blocks of the architecture of ReActNet, the proposed CNN for repetition detection. The colors denote layer types. Orange: convolution and non-linearity, red: max-pooling, blue: upsampling. The count of feature maps is shown under each convolution layer, and the size of the resulting feature maps is shown diagonally to the right of the layer (also visually as the size of the block). Finally, the blue sphere denotes addition.
\label{fig:architecture}}
\end{figure*}

\subsection{Data Augmentation}
\label{sec:augmentation}
In the case of a network that accepts visual input such as a regular color image, it is common practice to apply data augmentation in the form of geometric and intensity transformations~\cite{Krizhevsky} so as to force the neural network to become invariant to such transformations. Our case is different, since the goal is to learn the patterns generated from repetitive, periodic motion. The only useful form of data augmentation applicable to our input data would be one that achieves temporal scale invariance. This is effectively achieved by varying the size of the sub-blocks used during training.
The highlighted sub-blocks in Figure~\ref{fig:GT_subblocks} are of different size, following this data augmentation approach.

\subsection{Repetitive Activities Detection Network}
\label{sec:architecture}
{\noindent \bf ReActNet Architecture:} 
Given the defined input and target output, we propose, train, and evaluate ReActNet, a custom, lightweight convolutional neural network to learn this mapping. The employed architecture is a stack of hourglass modules~\cite{Newell2016}, that are essentially autoencoders~\cite{Hinton2011} with skip connections~\cite{He2016}.
The architecture consists of three stacked autoencoders. The reason we adopted an autoencoder-like network is because we are targeting an output with the same size as the input.
Furthermore, the autoencoder architecture offers good generalization by squeezing the information through the (spatially) low-resolution intermediate layers.

The proposed neural network is built using copies of two building blocks, an encoder and a decoder. This architecture is shown in Figure~\ref{fig:architecture}. The encoder first applies $16$ convolution filters of dimension $11 \times 11$. This is a rather large spatial dimension, however it is justified because the patterns we are looking for are rather large-scale and noisy.
The encoder then contains a ReLu activation layer, followed by a batch normalization layer~\cite{ioffe2015batch}. Another set of convolution, activation, and batch-norm layers follow in the original input dimension, and then a max pooling layer halves the resolution in each of the two spatial dimensions of the input. Two more sets of convolution, activation, and batch normalization follow, and in parallel to these layers, an identity residual connection~\cite{He2016} is also used. The skip connection is added to the result of the other branch, and a last max-pooling layer halves again the spatial resolution. In total, the encoder applies $5$ different convolution layers and two max pooling operations, resulting in output spatial resolution that is $1/4$ of the input resolution in each spatial dimension.
The decoder follows essentially the mirrored connectivity pattern of the encoder, upsampling its input by a factor of $4$ for each input spatial dimension. There are again $5$ layers applying a convolution operation, and a skip connection runs parallel to a group of two such convolutions.

\vspace*{0.2cm}
{\noindent \bf Loss Function:}
We build a classifier with two output classes so it is natural to use binary cross entropy (BCE)~\cite{Janocha2017} as the loss function for training:
\begin{equation}
    E(y, p) = -(y \log{p} + (1-y)\log(1-p)).
\label{eq:BCE}
\end{equation}
In Eq.(\ref{eq:BCE}), $y\in\{0,1\}$ is the ground truth class annotation, and $p\in[0,1]$ is the real-valued prediction of the network for this frame. In practice, we noticed that there is an imbalance between positive and negative examples in our training data. Specifically, our training set contains more segments with repetitive activities than non-repetitive ones.
This had an impact on the discriminative power of the trained models. To alleviate this problem, we used the weighted binary cross entropy (WBCE), a straightforward extension of the BCE loss:
\begin{equation}
    E(y, p) = -(w y \log{p} + (1-w)(1-y)\log(1-p)).
\label{eq:WBCE}
\end{equation}
In Eq.(\ref{eq:WBCE}), $y$ and $p$ are as in Eq.(\ref{eq:BCE}), and $w$ is a weight term balancing the two classes.
Using the BCE loss proved sufficient for our problem: given that we are essentially tackling a classification problem with only two classes, it was not needed to resort to more complex loss functions.
In practice it proved beneficial to apply intermediate supervision in each stage of the proposed architecture, anf to vary the weight parameter $w$ of WBCE in each stage. Specifically, the first stage used a small value for $w$, and the next two stages used increasingly larger values. This technique has the effect that both positive and negative examples are sufficiently weighted at some stage during training.

\vspace*{0.2cm}
{\noindent \bf Testing:}
At run-time, we adopt a sliding window approach. Given an input video, the distance matrix $M$ is formed. Then, a square window is centered around each point of its diagonal. This is fed to ReActNet which returns a square window of the same size with real-valued predictions. 
This sliding window approach results in multiple ($n_f$) predictions $p_i$, $1 \leq i \leq n_f$ per frame $f$, each in the range $[0,1]$, for the overlapping frames of consecutive windows. A frame $f$ is declared as being repetitive if: 
\begin{equation}
P_f = \sum_{i=1}^{n_f}\frac{p_i}{n_f}  > T. 
\label{eq:classify}
\end{equation}
In Eq.(\ref{eq:classify}), $T$ was set to $0.5$ in all experiments.

\section{Experimental Evaluation}
The experimental evaluation of the proposed method was performed on two recent  relevant datasets. A first category of experiments assessed the adopted design choices in an ablation study. Furthermore, the localization accuracy of ReActNet was compared to the current state of the art, for different choices of employed features. We also assessed the cross-dataset generalization of ReActNet by training it on a dataset and testing it on another.

\subsection{Training Details}
We implemented ReActNet using the Keras framework~\cite{chollet2015keras} on top of tensorflow~\cite{tensorflow2015-whitepaper}. The Adam optimizer was used to train it for $3$ epochs, with a learning rate value of $0.002$. For training, we employed an Nvidia GTX 1070 Ti GPU. On that machine, each epoch took $70$ seconds. The chosen range of sub-block sizes for data augmentation (see Section~\ref{sec:augmentation}) was between $100$ and $200$, with all the sub-blocks resized to $140 \times 140$ using Lanczos resampling.


\begin{figure*}[t]
\centering
\includegraphics[height=5cm]{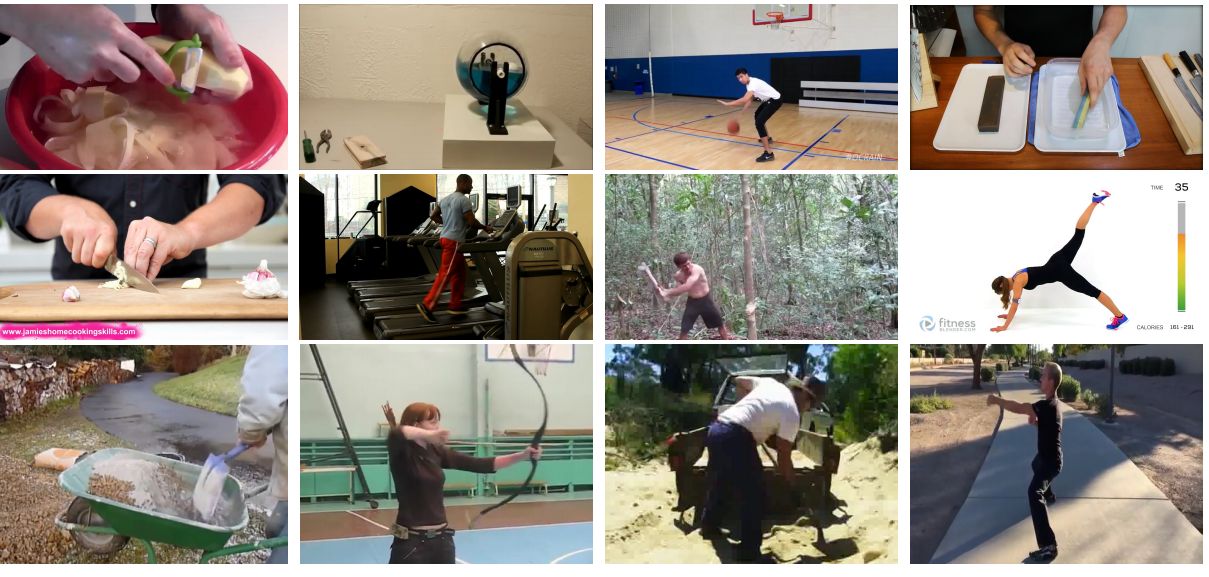} 
\hspace*{0.2cm}
\includegraphics[height=5cm]{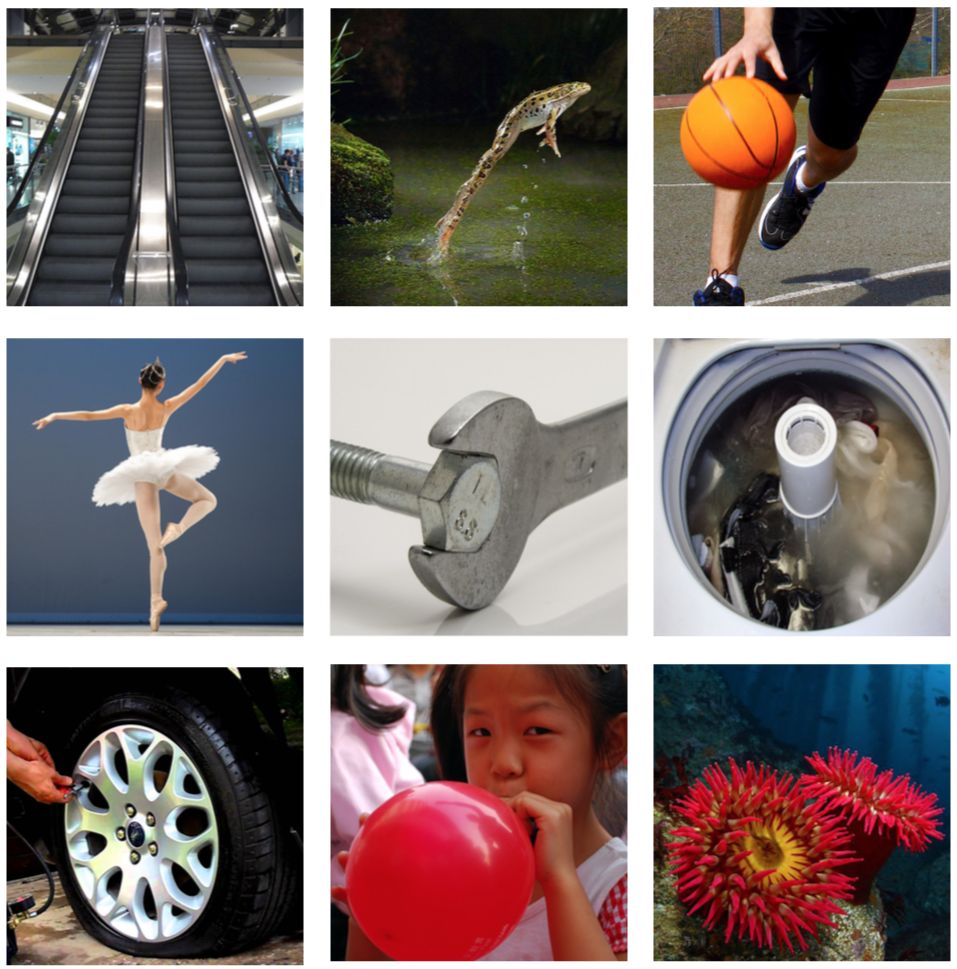} 
\caption{Video frames from the PERTUBE (left) and QUVA (right) datasets.
\label{fig:datasets}}
\end{figure*}


%

\subsection{Datasets}
For the evaluation of the proposed methodology, we use the PERTUBE~\cite{Panagiotakis} and the QUVA~\cite{Runia2018a} datasets (see Fig.~\ref{fig:datasets}).

\vspace*{0.2cm}
{\noindent \bf The PERTUBE dataset~\cite{Panagiotakis}:} This dataset\footnote{Available online at \url{https://www.ics.forth.gr/cvrl/pd/}} contains a set of videos depicting repetitive motions (human activities, object motions, etc) that were obtained from YouTube. There is a total of $50$ annotated videos in the dataset. Each frame of every video is annotated with information that denotes whether it belongs or not to a repetitive segment.

\vspace*{0.2cm}
{\noindent \bf The QUVA dataset~\cite{Runia2018a}:}
This dataset\footnote{Available online at \url{http://tomrunia.github.io/projects/repetition/}} is compiled for the related problem of repetition counting. It contains $100$ videos depicting human activities. Each video is appropriately annotated for repetition detection. 
Additionally to this information, specific frames are denoted as the starting frames of a new repetition, making the annotation useful for the task of repetition counting. We disregard this information, retaining only the repetitive/non-repetitive annotation per frame.

\subsection{Evaluation Metrics}
We view repetition detection as a classification problem. Thus, to evaluate the obtained results, we use the standard metrics of recall $\cal{R}$, precision $\cal{P}$, $F_1$ score and Overlap $\cal{O}$.
In our problem, precision quantifies how many of the frames that were classified as belonging to a repetitive segment are truly such. Recall quantifies the percentage of frames that actually belong to repetitive segments and were correctly classified by a method.

\subsection{Ablative Study}
In a first set of experiments, we justified specific design choices of ReActNet based on the PERTUBE dataset. The results of these experiments are presented in Table~\ref{table:ablation}. In each column, the best result is highlighted with bold.
The results were obtained by performing five randomized repetitions with half the videos of the dataset used as training and the other half as test. Then, the reported value is the average of the performance on the test set in the five randomized repetitions.
The first line of Table~\ref{table:ablation} refers to the architecture as this has been described in Section~\ref{sec:architecture}. 
The experiment ``Filter size 11 $\times$ 11'' tests a larger filter size (11 $\times$ 11) for the first convolutional layer of the autoencoders, compared to the 5 $\times$ 5 of the proposed method.
The ``Single autoencoder'' experiment uses a single autoencoder instead of the three stacked autoencoders of the proposed method. The experiment ``Without skip connection'' disables the ResNet-like skip connections of the proposed method. Finally, the ``Intermediate supervision'' experiment explores the idea of adding two extra terms in the loss function of the network, at the end of each autoencoder in the stack. The target is the same for all three loss terms, the ground truth classification of the sub-block. Evidently, the proposed method outperforms all other network variants.
\begin{table}[t]
\caption{Meta-parameter study of the proposed neural network. Along with the values of the metrics, the standard deviations over the repeated experiments are also shown.}
\begin{center}
\scalebox{0.65}{
\vspace*{-10mm}
\begin{tabular}{|l|c|c|c|c|}
\hline 
\cal{Configurations}                 & $\cal{R} (\%)$ &           $\cal{P} (\%)$ & $F_1 (\%)$  & $\cal{O} (\%)$ \tabularnewline
\hline\hline 
\hline 
 ReActNet (proposed)                 & $\textbf{87.6}\pm 3.4$   & $ 85.1\pm1 $       & $ \textbf{85.8}\pm2.2 $ & $ \textbf{75.5}\pm3.9$ \tabularnewline
\hline 
 Filter size 11$\times$11            & $ 86.6\pm9.2 $  &          $  85.2\pm3.9 $    & $ 85.3\pm3.2 $          & $ 75.6\pm3.5 $ \tabularnewline 
\hline 
 Single autoencoder                  & $80.7\pm4    $  &          $ 84.0\pm3.1 $     & $ 81.6\pm3.1 $          & $ 69.5\pm3.4 $ \tabularnewline 
\hline 
Without skip conn.                   & $ 84.6\pm7.6 $  &          $ 83.7\pm4.7 $     & $ 83.6\pm2.6 $          & $ 73.0\pm3.3 $  \tabularnewline 
\hline
Without interm. supervision          & $ 84.2\pm6.2 $  &          $\textbf{86.5}\pm1.9 $  & $ 85.3\pm3.7 $     & $ 75.0\pm4.7 $ \tabularnewline 
\hline 
\hline 
\end{tabular}
}
\label{table:ablation}
\end{center}
\end{table}
In the following, unless otherwise stated, ReActNet and the term ``proposed method" refer to the architecture presented in Section~\ref{sec:architecture} and evaluated in the first row of Table~\ref{table:ablation}.

\begin{table}[t]
\caption{Comparative evaluation of ReActNet with~\cite{Panagiotakis} on the PERTUBE dataset with CNN-based (VGG19) and hand-crafted (IDT) features. Training of the proposed method has been performed on the PERTUBE dataset.}
\begin{center}
\scalebox{0.75}{
\vspace*{-10mm}
\begin{tabular}{|l|l|c|c|c|c|}
\hline 
\cal{Method} & \cal{Features} & $\cal{R} (\%)$ & $\cal{P} (\%)$ & $F_1 (\%)$  & $\cal{O} (\%)$ \tabularnewline
\hline\hline 
\cite{Panagiotakis}   & IDT    & 84.1 & 75.7 & 77.0 & 67.7\tabularnewline 
\hline 
ReActNet &    IDT           & $87.1\pm8.4$ & $78.6\pm5.7$ & $82.0\pm4.6$ & $471.9\pm4$ \tabularnewline
\hline 
\cite{Panagiotakis} &  VGG19    & 79.2 & 68.1 & 71.1 & 61.8\tabularnewline 
\hline 
ReActNet &  VGG19    & $\textbf{87.6}\pm 3.4$ & $\textbf{85.1}\pm1$ & $\textbf{85.8}\pm2.2$ & $\textbf{75.5}\pm3.9$ \tabularnewline
\hline 
\hline 
\end{tabular}}
\label{table:pancomparison}
\end{center}
\end{table}

\subsection{Comparison to SoA \& the Impact of Features}
The recent method by Panagiotakis et al.~\cite{Panagiotakis} solves the problem of repetition localization by using a distance matrix computed on an Improved Dense Trajectories (IDT)~\cite{Wang2013} representation of each frame of the sequence. This distance matrix is then appropriately manipulated to extract the repetitive parts of the video. We note that both the method in~\cite{Panagiotakis} and the proposed method consist of a first phase that builds a distance matrix based on some feature representation of the frames of a video (IDT features for~\cite{Panagiotakis}, VGG19 features for us) and a second phase that localizes repetitions on this distance matrix (hand-crafted filtering process followed by Discrete Time Warping for~\cite{Panagiotakis}, ReActNet for us). It is therefore possible to interchange parts of the two methodologies, assessing the impact of deep neural networks on the proposed approach in a total of $4$ different experiments.
%
%
Table~\ref{table:pancomparison} presents the results we obtained with the $4$ different variants on the PERTUBE dataset. The first column regards the employed method (\cite{Panagiotakis} and ReActNet). The second column regards the employed features (Improved Dense Trajectories~\cite{Wang2013} suggested by~\cite{Panagiotakis}, VGG19 features~\cite{Krizhevsky} suggested in this work). 

For the approach in~\cite{Panagiotakis}, the results were obtained by a single run over the whole dataset, using the publicly available implementation provided by the authors of that work\footnote{The implementation of the method in~\cite{Panagiotakis} is available at \url{https://sites.google.com/site/costaspanagiotakis/research/pd}}.
For our approach, the values were estimated with same methodology as above, using five repetitions with randomized training and test sets.
For each of the employed metrics, the best result is highlighted in bold. It can be verified that the proposed approach achieves the best results compared to all the other possible configurations. 
Interestingly, the worst of our two variants performs better than the best of the variants of~\cite{Panagiotakis}, that is, regardless of whether we use IDT or VGG19 features. Thus, the quality of the obtained results is attributed mostly to the method itself and to a lesser extend to the employed features. 

\subsection{Cross-dataset Validation}
We performed a set of experiments to investigate how ReActNet generalizes across datasets. In a first experiment we trained the proposed method on the PERTUBE dataset, and assessed its performance on the QUVA dataset. In a second experiment, we swapped the training and test sets.

\begin{table}[t]
\caption{Cross-dataset evaluation: The proposed method, ReActNet, was trained on  QUVA and evaluated on PERTUBE.}
\begin{center}
\scalebox{0.95}{
\begin{tabular}{|l|c|c|c|c|c|}
\hline 
\cal{Methods} & \cal{Features} & $\cal{R} (\%)$ & $\cal{P} (\%)$ & $F_1 (\%)$  & $\cal{O} (\%)$ \tabularnewline
\hline\hline 
 ReActNet &VGG19 & 82.8 & 76.8 & 79.7 & 67.8\tabularnewline 
\hline 
\hline 
\end{tabular}}
\label{table:quva2per}
\end{center}
\end{table}

\begin{table}[t]
\caption{Cross-dataset evaluation: The proposed method was trained on PERTUBE and evaluated on  
QUVA. The method in~\cite{Panagiotakis} requires no training.}
\begin{center}
\scalebox{0.95}{
\begin{tabular}{|l|c|c|c|c|c|}
\hline 
\cal{Methods} & \cal{Features} & $\cal{R} (\%)$ & $\cal{P} (\%)$ & $F_1 (\%)$    & $\cal{O} (\%)$ \tabularnewline
\hline\hline 
\hline 
 ReActNet     & VGG19          & \textbf{89.5}  & \textbf{87.4}  & \textbf{88.5} & \textbf{80.3} \tabularnewline
\hline 
\cite{Panagiotakis}     & VGG19          & 83.8           & 80.6           & 83.1          & 72.8           \tabularnewline 
\hline 
\hline 
\end{tabular}}
\label{table:per2quva}
\end{center}
\end{table}

\vspace*{0.2cm}
{\noindent \bf Training on QUVA, testing on PERTUBE:}
Table \ref{table:quva2per} shows the performance of ReActNet when trained on QUVA and tested on PERTUBE. The obtained results demonstrate that there is a small performance drop compared to training on PERTUBE and testing on PERTUBE (comparison with the fourth line of Table~\ref{table:pancomparison}). The performance drop is attributed to the smaller diversity of QUVA compared to PERTUBE. Still, the proposed approach generalizes well.

\vspace*{0.2cm}
{\noindent \bf Training on PERTUBE, testing on QUVA:}
The results of this experiment are shown in Table~\ref{table:per2quva}. 
When training on PERTUBE, the results of testing on QUVA are better than those of testing on PERTUBE (Table~\ref{table:pancomparison}, fourth row). Given that training involved the same subset of PERTUBE in both cases, this serves as a further indication that PERTUBE is more diverse than QUVA (see previous paragraph). For comparison, the second row of Table~\ref{table:per2quva} shows the performance of~\cite{Panagiotakis} on the QUVA dataset using our proposed deep features. It can be verified the the proposed approach outperforms~\cite{Panagiotakis} in all performance metrics and that there is a $18\%$ increase in overlap ${\cal O}$.

\subsection{Qualitative Results}
\begin{figure}[t]
\centering
\includegraphics[width=0.44\columnwidth]{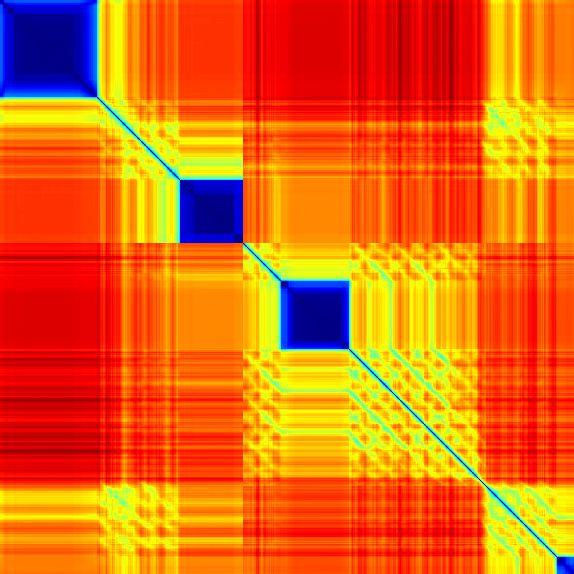}
\hspace*{0.2cm}
\includegraphics[width=0.44\columnwidth]{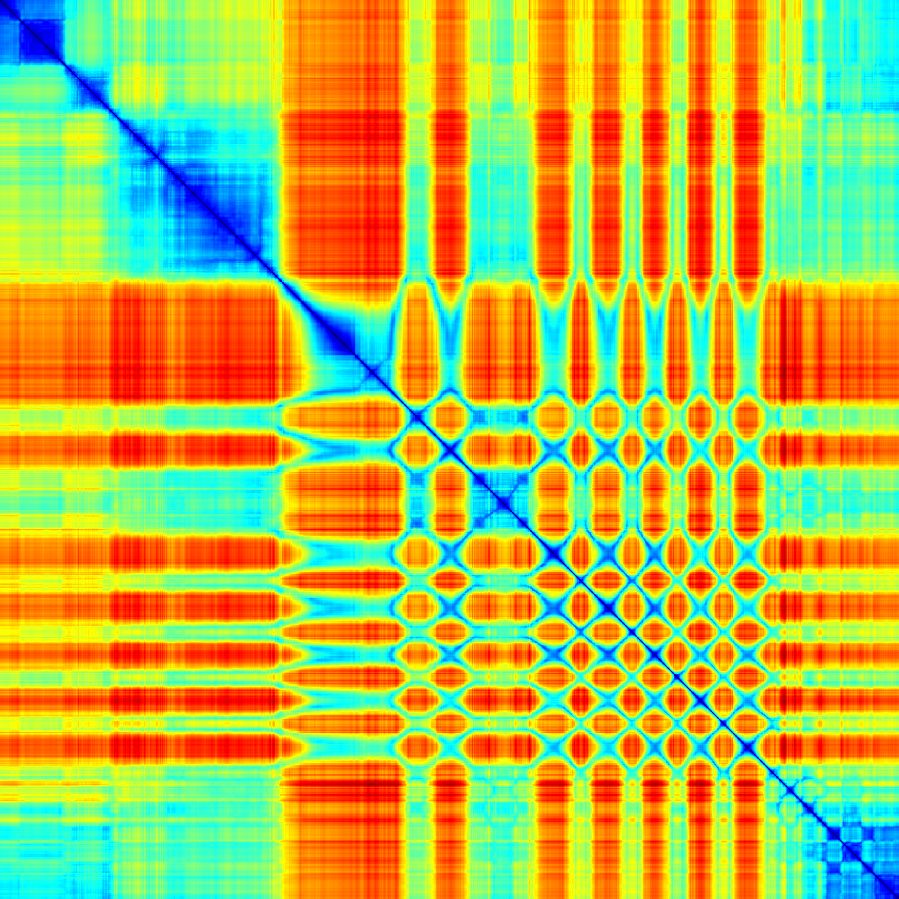}\\
\includegraphics[width=0.44\columnwidth,height=0.5cm]{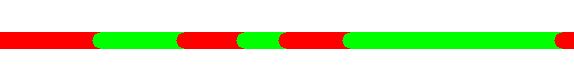} 
\hspace*{0.2cm}
\includegraphics[width=0.44\columnwidth,height=0.5cm]{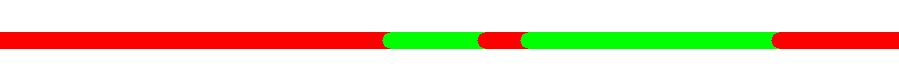}\\
\includegraphics[width=0.44\columnwidth,height=0.5cm]{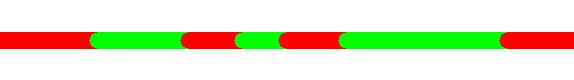} 
\hspace*{0.2cm}
\includegraphics[width=0.44\columnwidth,height=0.5cm]{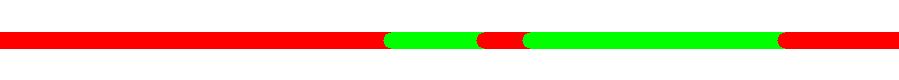}\\
\vspace*{0.3cm}
\includegraphics[width=0.44\columnwidth]{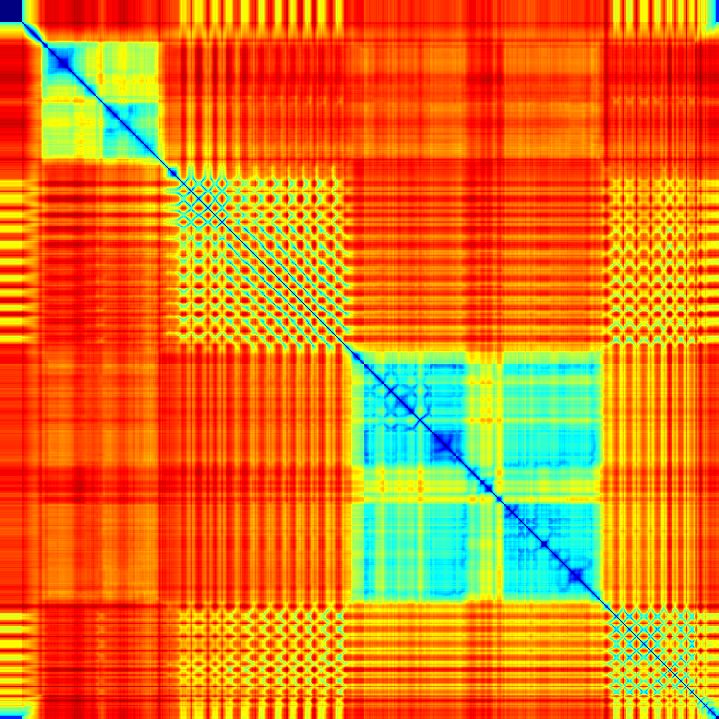}
\hspace*{0.2cm}
\includegraphics[width=0.44\columnwidth]{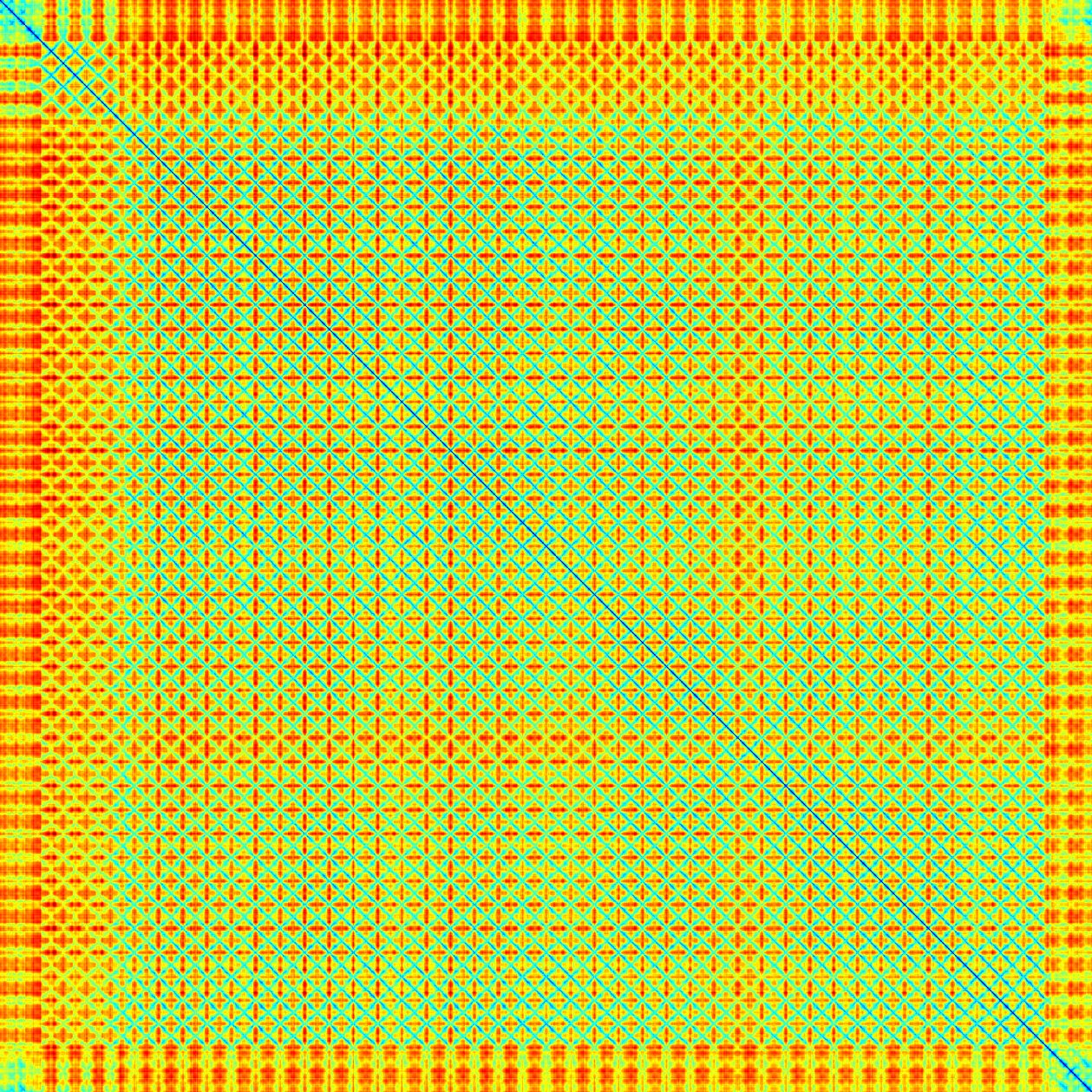}\\
\includegraphics[width=0.44\columnwidth,height=0.5cm]{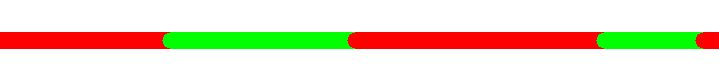} 
\hspace*{0.2cm}
\includegraphics[width=0.44\columnwidth,height=0.5cm]{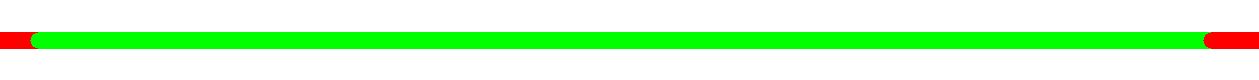}\\
\includegraphics[width=0.44\columnwidth,height=0.5cm]{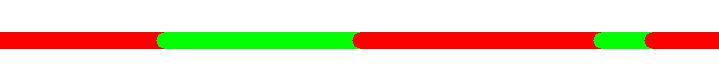} 
\hspace*{0.2cm}
\includegraphics[width=0.44\columnwidth,height=0.5cm]{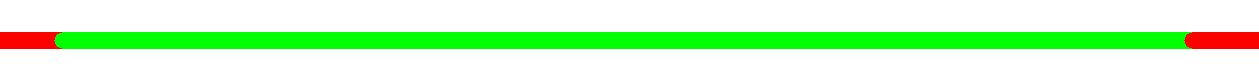}\\
\vspace*{0.3cm}
\includegraphics[width=0.44\columnwidth]{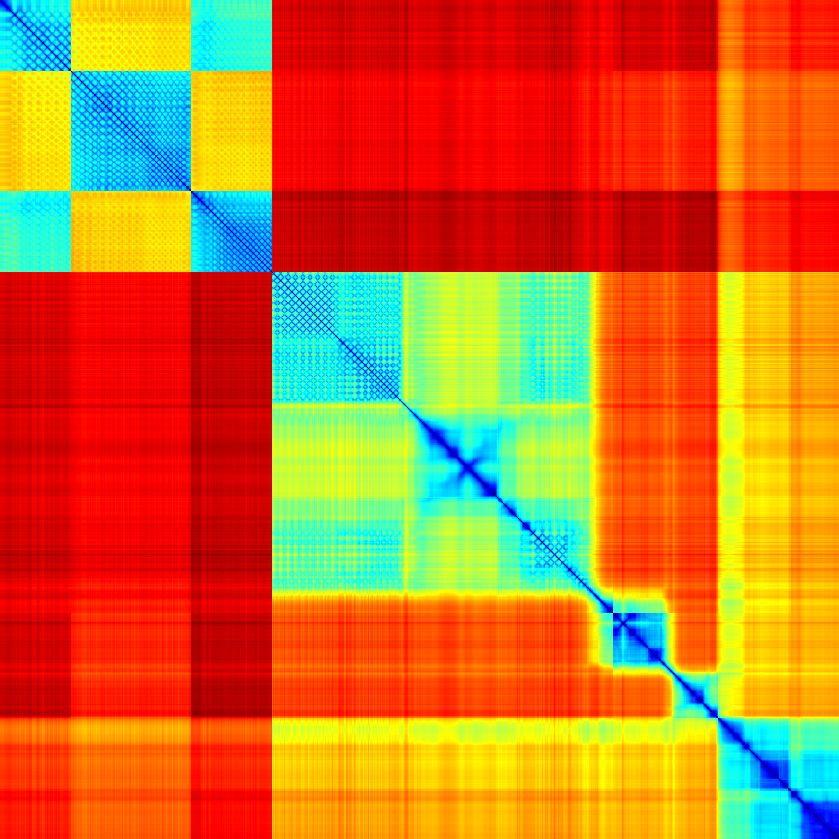}
\hspace*{0.2cm}
\includegraphics[width=0.44\columnwidth]{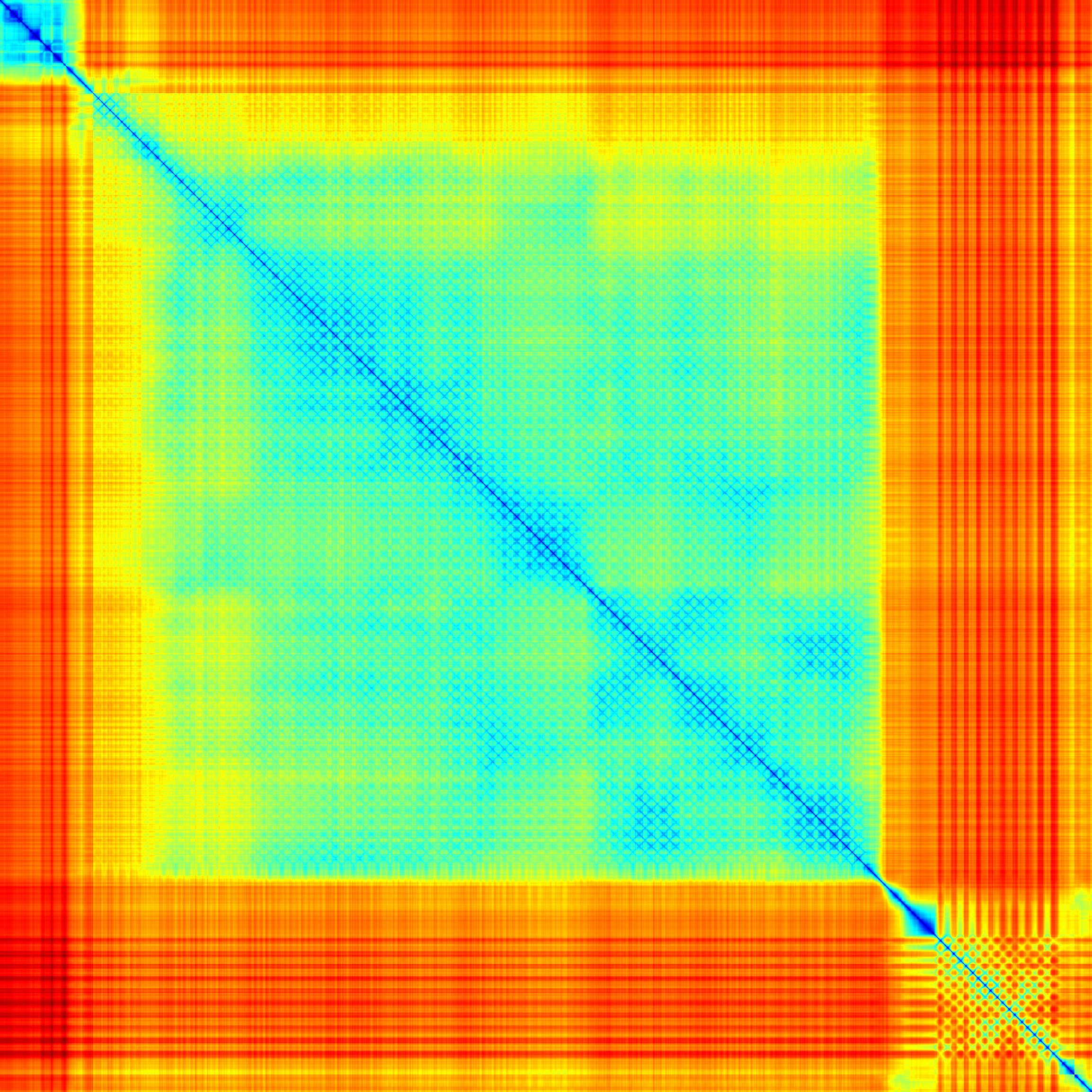}\\
\includegraphics[width=0.44\columnwidth,height=0.5cm]{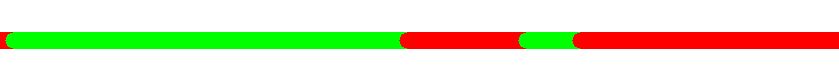} 
\hspace*{0.2cm}
\includegraphics[width=0.44\columnwidth,height=0.5cm]{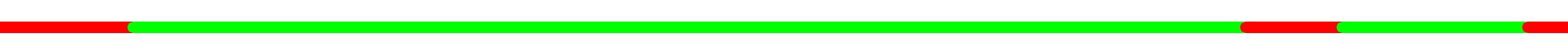}\\
\includegraphics[width=0.44\columnwidth,height=0.5cm]{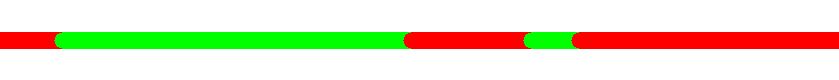} 
\hspace*{0.2cm}
\includegraphics[width=0.44\columnwidth,height=0.5cm]{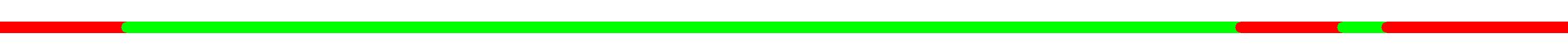}\\
\vspace*{0.3cm}
\includegraphics[width=0.44\columnwidth]{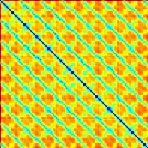}
\hspace*{0.2cm}
\includegraphics[width=0.44\columnwidth]{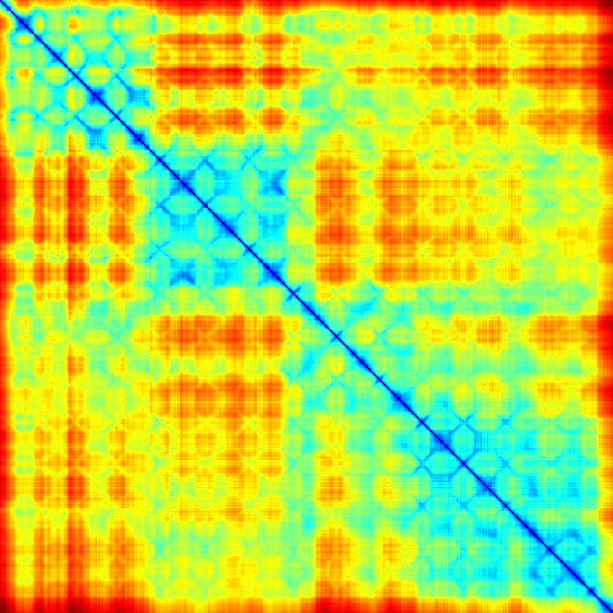}\\
\includegraphics[width=0.44\columnwidth,height=0.5cm]{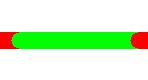} 
\hspace*{0.2cm}
\includegraphics[width=0.44\columnwidth,height=0.5cm]{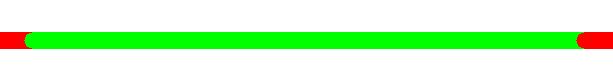}\\
\includegraphics[width=0.44\columnwidth,height=0.5cm]{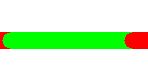} 
\hspace*{0.2cm}
\includegraphics[width=0.44\columnwidth,height=0.5cm]{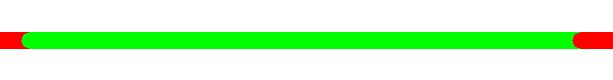}
\caption{The distance matrices, the repetition localization estimation (1st green/red bar) and the ground truth (2nd green/red bar) for six sequences of the PERTUBE (top three rows) and the QUVA (last row) datasets.
\label{fig:results}}
\end{figure}

Figure~\ref{fig:results} shows the distance matrices, the repetition localization estimation (1st green/red bar) and the ground truth (2nd green/red bar) for six sequences of the PERTUBE (top three rows) and the QUVA (bottom row) datasets. The repetition localization results are better illustrated in the supplementary material\footnote{Available online at \url{https://youtu.be/pPqg1lMkuaQ}} of this paper. More sequences have been selected from  PERTUBE because it contains more complex activities of multiple repetitive segments per video.

\section{Conclusions}
We presented a novel method for the temporal localization of repetitive activities within a video. In each frame of the video, the method extracts features using a deep neural network and then estimates a matrix of pairwise distances between video frames. Then, ReActNet, a specially designed convolutional neural network classifies the frames of the video as belonging to repetitive segments or not. Extensive experiments backed the design choices behind the proposed method, verified the use of the employed features, compared the approach to a state of the art method and assessed the generalization potential across datasets.
%
From a computational point of view, a shortcoming of the proposed method is that the current feature extraction method prevents its use in real-time or online applications.
%
Future plans include the investigation of alternatives that may allow online or even real-time operation, an extension of the existing framework to tackle the problem of repetition counting, and mechanisms for performing localization of the video periodicity also in the spatial domain. 


\section*{Acknowledgments}
This work was partially supported by the H2020-ICT-2016-1-731869 project Co4Robots.

{\small
\bibliographystyle{ieee}
\bibliography{egbib}
}

\end{document}